\documentclass[letterpaper]{article}
\usepackage{aaai}
\usepackage{times}
\usepackage{helvet}
\usepackage{courier}
\usepackage{natbib}

\usepackage{amsmath}
\DeclareMathOperator*{\argmax}{arg\,max}
\DeclareMathOperator*{\argmin}{arg\,min}

\usepackage[linesnumbered]{algorithm2e}
\usepackage{algorithmic}
\usepackage{graphicx}
\usepackage{url}
\usepackage{tabularx}
\usepackage{booktabs}
\usepackage{multirow}
\usepackage{CJK}

\begin{document}
% Kun Xu, Linfeng Song .....
\title{Coordinated Reasoning for Cross-Lingual Knowledge Graph Alignment}
\author{Kun Xu$^{1}$, Linfeng Song$^{1}$, Yansong Feng$^{2}$, Yan Song$^{3}$, Dong Yu$^{1}$\\
$^{1}$Tencent AI Lab, Seattle, US\\
$^{2}$Peking University, Beijing, China\\
$^{3}$Sinovation Ventures \\
% \{xukun, evancheung, fengyansong, zhaodongyan\}@pku.edu.cn,  huangsf@cn.ibm.com\\
}
\maketitle
\begin{CJK*}{UTF8}{gbsn}
\begin{abstract}
\begin{quote}
Existing entity alignment methods mainly vary on the choices of encoding the knowledge graph, but they typically use the same decoding method, which independently chooses the local optimal match for each source entity. This decoding method may not only cause the ``many-to-one'' problem but also neglect the coordinated nature of this task, that is, each alignment decision may highly correlate to the other decisions.
In this paper, we introduce two coordinated reasoning methods, i.e., the Easy-to-Hard decoding strategy and joint entity alignment algorithm.
Specifically, the Easy-to-Hard strategy first retrieves the model-confident alignments from the predicted results and then incorporates them as additional knowledge to resolve the remaining model-uncertain alignments. To achieve this, we further propose an enhanced alignment model that is built on the current state-of-the-art baseline. In addition, to address the many-to-one problem, we propose to jointly predict entity alignments so that the one-to-one constraint can be naturally incorporated into the alignment prediction. Experimental results show that our model achieves the state-of-the-art performance and our reasoning methods can also significantly improve existing baselines.
\end{quote}
\end{abstract}

\section{Introduction}

%% importance of KG
Knowledge graphs (KGs), such as Freebase \citep{bollacker2008freebase} and DBpedia \citep{auer2007dbpedia},
represent world-level factoid information of entities and their relations in a graph-based format.
They have been successfully used in many natural language processing applications, such as question answering \citep{berant2013semantic,bao2014knowledge,yih2015semantic,kun_question_2016,DBLP:journals/corr/DasZRM17}
and relation extraction \citep{mintz2009distant,hoffmann2011knowledge,min2013distant,zeng2015distant}.
To date, there have been many KGs in different languages, with each being created in one language \citep{franco2016systematic}. They share lots of the same facts, and each also provides rich additional information that the others do not cover.
Thus, it is very beneficial to establish the cross-lingual alignments between KGs, so that the combined KG can provide richer knowledge for downstream tasks.
Therefore, the cross-lingual KG alignment task, which automatically matches entities between multilingual KGs, is proposed to address this problem.

%% existing-method overview
Most recently, several approaches based on cross-lingual entity embeddings \citep{hao2016joint,chen2017multilingual,sun2018cross} or graph neural networks \citep{wang2018cross,xu2019cross,wu2019relation} have been proposed for this task.
In particular, \citet{xu2019cross} introduces the topic entity graph to capture the local context information of an entity within the KG, and further tackles this task as a graph matching problem by proposing a graph matching network. This work significantly advanced the state-of-the-art accuracies across several datasets.

%dramatically advanced the state-of-the-art accuracies by 25.0+ absolute points across several datasets, showing that free MT service and local context can alleviate the cross-lingual issue and other ambiguities.
% convert foreign KG nodes into English with Google translate\footnote{https://translate.google.com}, before leveraging a graph convolutional network (GCN) to represent local contexts of associated entities for further disambiguation.

%% their first shortcoming
Despite the excitingly progressive results that have been shown, all previous works fail to consider the coordinated nature of this task, that is, each alignment decision may highly correlate to the other decisions.
For example, all existing models independently align each source entity, which may result in the many-to-one mapping, i.e.,
more than one source entities are aligned to the same target entity.
In particular, we analyze the results of \citet{xu2019cross} and find that nearly $8$\% of the alignments are many-to-one mappings.
One intuitive solution is to align these entities in a greedy fashion,
that is, assign one alignment at each time with a constraint that all alignments are one-to-one mappings. 
However, this may introduce the error propagation, since
each decision error may propagate to the future decisions.
On the other hand, given the fact that the KGs are large,
it is also impractical to jointly assign all alignments, due to the massive search space.

\begin{figure}
    \centering
    \includegraphics[width=1.0\linewidth]{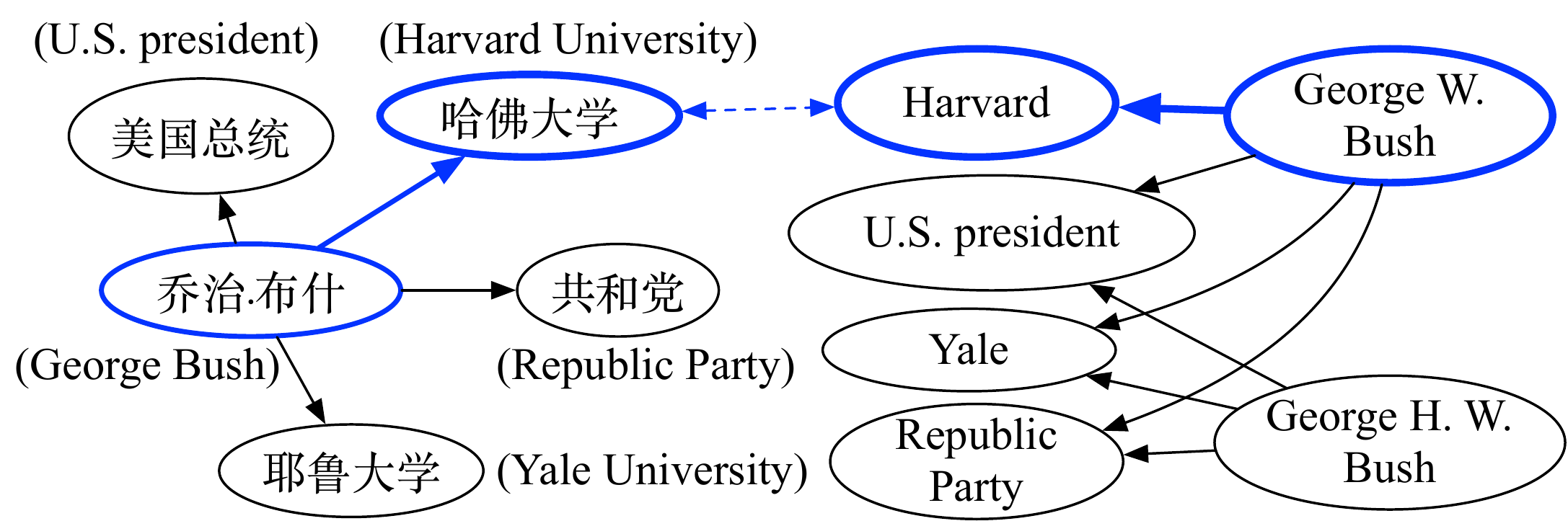}
    \caption{A challenging entity matching example.}
    \label{fig:easy_to_hard_decode}
    \vspace{-3mm}
\end{figure}

We analyze the results of existing alignment baselines and find the second type of errors are caused by the existence of \textbf{\textit{adversarial}} entities that have similar surface strings and KG neighbors with the ground truth.
It is challenging for existing approaches to disambiguate these entities since previous methods mainly rely on the embeddings that are
derived by encoding the surface strings and KG neighbors.
Figure~\ref{fig:easy_to_hard_decode} gives such an example, where it is ambivalent for a model to align {\small{乔治$\cdot$布什}} (George Bush) to ``George W. Bush'' or ``George H. W. Bush'', because both candidates have similar surface strings and share several common neighbors (such as ``Republic Party'' and ``U.S. president'').
% This type of errors are also significant, which we will show later in our experiments.

%% our method
In this paper, we propose to alleviate these two types of errors using 
two coordinated reasoning methods, i.e., the Easy-to-Hard strategy and joint entity alignment algorithm.
Specifically, the Easy-to-Hard strategy leverages an iterative approach, where the most model-confident (\textbf{easy}) alignments predicted in the previous iteration are provided as additional inputs to the current iteration for resolving the remaining model-uncertain (\textbf{hard}) alignments.
This idea is motivated by our observation that the model-confident alignments are mostly correct, and thus they can provide reliable clues for other decisions with less model confidence.

To address the many-to-one problem, we propose a joint entity alignment algorithm that finds the global optimal entity alignments that satisfy the one-to-one constraint.
This problem is essentially a fundamental combinatorial optimization problem whose exact solution can be found by the Hungarian algorithm \citep{kuhn1955hungarian}.
However, since this algorithm takes a high time complexity of $O(N^4)$ for KGs of $N$ nodes, it is impractical to apply this algorithm in our framework directly.
To address this, we propose a simple yet effective solution that breaks down the whole search space into small isolated pieces, so that each piece could be efficiently solved with the Hungarian algorithm.
Experiments on the benchmark datasets show that our proposed coordinated reasoning methods can not only improve the current state-of-the-art performance but also significantly boost the performance of previous approaches.
% Our code is available at {\small \url{http://github.com/anonymous}}.

\section{Related Work}
Our work is mainly related to two lines of research:
network embedding and knowledge graph alignment.
\subsection{Graph Convolutional Networks}
Recently, there has been an increasing interest in extending neural networks to deal with graphs.
\cite{defferrard2016convolutional} proposed a spectral graph theoretical formulation of CNNs on graphs and a convolutional network extending the conventional CNNs to non-Euclidean space.
\cite{kipf2016semi} further extended this idea and proposed graph convolutional neural networks (GCNs) to integrate the connectivity patterns and feature attributes of graph-structured data,
and achieved decent results in semi-supervised classification.
Thereafter, a series of improvements and extensions were proposed based on GCN. GAT \citep{velivckovic2017graph} employs the attention mechanism to GCNs, in which each node gets an importance score based on its neighborhood, thus providing more expressive representations for nodes.
Furthermore, the R-GCNs \citep{schlichtkrull2018modeling} are proposed to model relational data and have been successfully exploited in link prediction and entity classification.
Inspired by the capability of GCNs on learning node representations, we employ the GCN to build our entity alignment framework.

\subsection{Entity Alignment}
Earliest approaches of entity alignment usually require expensive expert efforts to design model features \citep{mahdisoltani2013yago3}. Recently, embedding based methods have been proposed to address this issue.
MTransE \citep{chen2017multilingual} employs TransE \citep{bordes2013translating} to embed entities and relations of each knowledge graph in a separate space, and then provides five different variants of transformation functions to project the embedded vectors from one subspace to another. The candidate set of one entity's correspondence in the other knowledge graph can be obtained by ranking the distance between them in the transformed space.
ITransE \citep{zhu2017iterative} utilizes TransE to learn one common low-dimensional subspace for all knowledge graphs, with the constraint that the observed anchor seeds from different knowledge graphs share the same vector representation in the subspace.
AlignE \citep{sun2018cross} also adopts TransE to learn network embeddings, and applies parameter swapping to encode network into a unified space. NTAM \citep{li2018non} utilizes a probabilistic model for the alignment task.
Instead of using TransE to derive entity embeddings from the knowledge graph, various GCN based methods
\citep{wang2018cross,DBLP:conf/ijcai/Ye0FZW19,wu2019relation} that use the conventional GCN to encode the entities and relations have been proposed to perform the alignment.
Different with those methods that still follow previous works that rely on learned entity embeddings to rank alignments, \cite{xu2019cross} views this task as a graph matching problem and further proposes a graph matching neural network that additionally considers the matching information of an entity's neighborhood to perform the prediction.

Despite these approaches achieve progressive results, all current works focus on encoding the entities and relations, while neglecting the fact that the decoding strategy may have a considerable impact over the final performance.
In this paper, we explore the coordinated nature of this task and propose two types of reasoning methods to improve the performance of these baselines.

% It views the relations as influential weights on entities in proportion to the number of entities connected with different types of relations.

\section{Problem Formulation}
Formally, a KG is represented as $G = (E, R, T)$,
where $E$, $R$, $T$ are the sets of entities, relations, and triples, respectively.
Let $G_{1} = (E_{1}, R_{1}, T_{1})$ and $G_{2} = (E_{2}, R_{2}, T_{2})$ be two heterogeneous KGs to be aligned.
That is, an entity in $G_{1}$ (source entity) may have its counterpart in $G_{2}$ (target entity) in a different language or different surface names.
As a starting point, we can collect a small number of equivalent entity pairs between $G_{1}$ and $G_{2}$ as the alignment seeds.
We define the entity alignment task as automatically finding more equivalent entities using the alignment seeds as training data.

\section{Coordinated Reasoning}
All existing works follow the conventional framework that first encodes the context information of the source entity within the KG into a distributional representation and then ranks the candidate target entities according to the representation similarities.
These works may differ in the choice of the encoder, such as TransE or GCN, but all of them utilize the same decoding method, which simply picks the local optimal candidate for each source entity without considering the global alignment coherence.
For example, more than one source entities may be aligned to the same target entity, causing the \textit{many-to-one} problem.
This simple decoding strategy also neglects the coordinated nature of this task, that is, previously predicted alignments are also helpful to future predictions.

Motivated by these observations, we propose two types of coordinated reasoning methods.
First, to address the many-to-one problem, we jointly predict alignments by explicitly incorporating the one-to-one constraint into the decoding.
Second, we propose a new Easy-to-Hard decoding strategy that
first resolves the most model-confident alignments and then uses them as additional evidence to better handle the model-uncertain alignments.

\subsection{Easy-to-Hard Decoding}
All existing models independently predict alignments for source entities while neglecting the fact that the decoding strategy may have a significant impact over the performance.
Figure~\ref{fig:easy_to_hard_decode} illustrates such an example where the goal is to align {\small{乔治$\cdot$布什}} (George Bush) from the Chinese KG into the English KG.
Given its two candidates, i.e., \textit{George W. Bush} and \textit{George H. W. Bush}, it is challenging for previous methods to find the correct alignment (\textit{George W. Bush}) since these candidates have almost the same neighbors, except that George W. Bush graduated from Harvard University while George H. W. Bush not. On the other hand, we can see that the Chinese KG includes a fact, {$<$\small{乔治$\cdot$布什}} graduated from {\small{哈佛大学}} (Harvard University)$>$, which is strong evidence for aligning {\small{乔治$\cdot$布什}} to George W. Bush.
Intuitively, if a model could first align {\small{哈佛大学}} to the Harvard University and introduce this as additional knowledge, it could be more easy for the model to find the correct alignment for {\small{乔治$\cdot$布什}}.
Compared to the alignment for {\small{乔治$\cdot$布什}}, which is \textbf{H}ard to resolve, the alignment for {\small{哈佛大学}} is relatively \textbf{E}asier.

\begin{figure*}[ht!]
    \centering
    \includegraphics[width=1.0\linewidth]{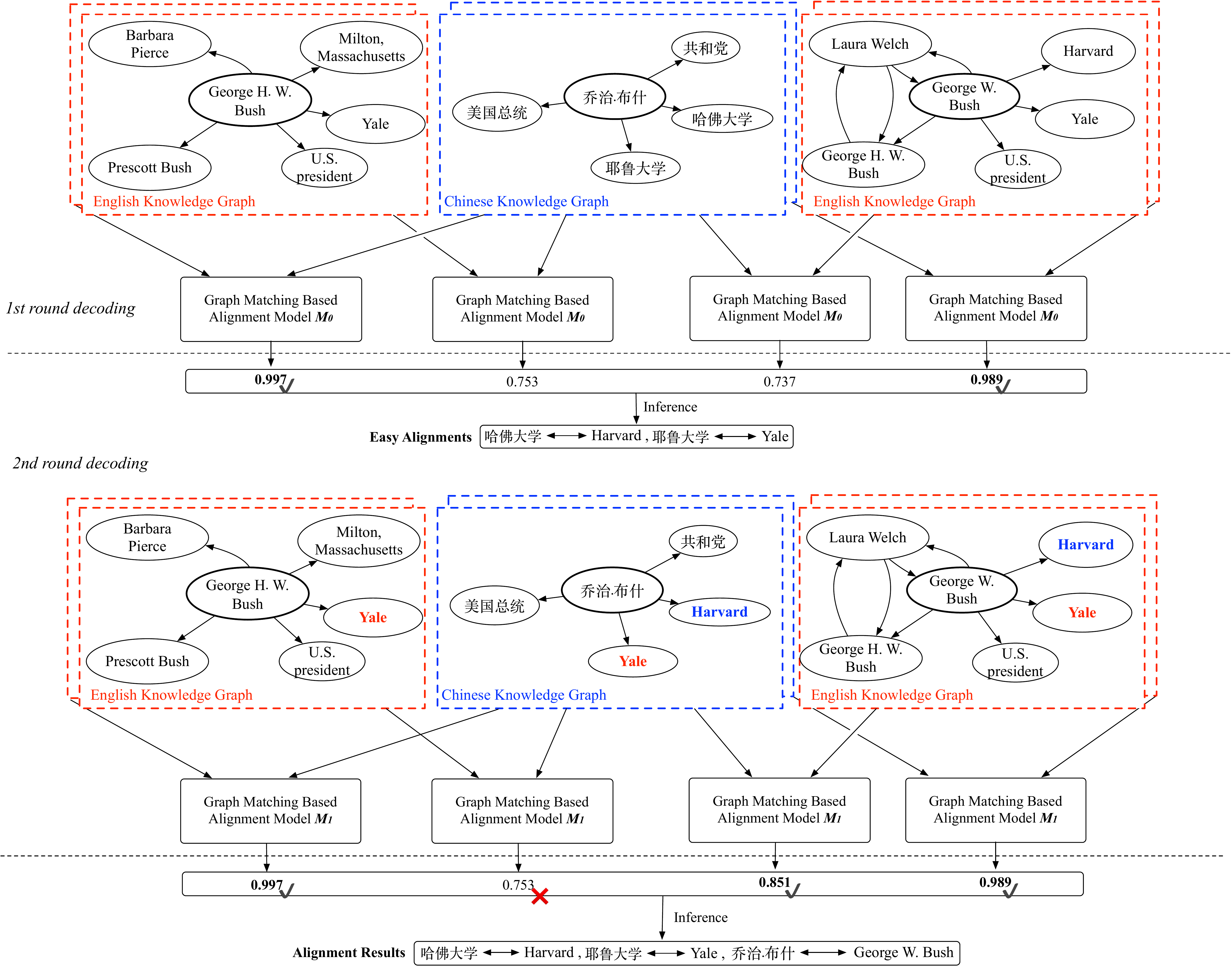}
    \caption{A running example of our Easy-to-Hard decoding strategy for aligning \textit{George Bush} in the English and Chinese knowledge graph. After the first round decoding, the baseline model aligns {\small{哈佛大学}} to Harvard and {\small{耶鲁大学}} to Yale, because their probabilities predicted by \textit{M$_0$} is higher than $\alpha$. After introducing these information, our enhanced model \textit{M$_1$} increased the probability of aligning {\small{乔治$\cdot$布什}} to George W. Bush while decreasing the probability of its alignment to George H. W. Bush.}
    \label{fig:framework}
    \vspace{-3mm}
\end{figure*}

Inspired by the above observation, in this paper, we propose a new
decoding method, namely Easy-to-Hard strategy, which first attempts to resolve ``easy'' alignments in the test set and then incorporates them as additional knowledge into the model to better tackle the remaining ``hard'' alignments.
There are two main challenges here.
First of all, it is difficult to determine whether an alignment is easy or hard to resolve.
Second, existing dominant models are mainly built on the graph neural networks, and it is unclear how to integrate such additional knowledge into their models.

We analyze the alignment results of three baseline methods, i.e., \cite{wang2018cross}, \cite{xu2019cross} and \cite{wu2019relation}.
Interestingly, we find that all these baselines could achieve at least $99.5$\% accuracy for those alignments with normalized probabilities over $0.9$. This result is coherent with our expectation since a higher probability typically suggests that the model is more confident about the prediction and also indicates that this alignment is easier for the model to resolve.
Therefore, we apply the following steps to decode the test set iteratively.

\vspace{0.2em}
\begin{center}
\begin{tabularx}{0.95\linewidth}{cX}
    \toprule
    Step & Description \\
    \midrule
    1 & Employ an alignment model to predict alignments for all source entities in the test set. \\
    2 & Use a predefined probability threshold $\alpha$ to refine those alignments. In particular, assignments with probabilities \textit{higher than} $\alpha$ are regarded as \textbf{easy} alignments while the others are viewed as \textbf{hard} alignments. \\
    3 & If more than $K$ easy alignments are found in Step 2, take these easy alignments as additional knowledge and incorporate them into the alignment model to establish alignments for the remaining entities (go to Step 1); otherwise, return all alignments.\\
    \bottomrule
\end{tabularx}
\end{center}
\vspace{0.2em}
After establishing easy assignments in each decoding step,
we need to incorporate them as additional knowledge into the alignment model for the next round decoding.
This design heavily depends on alignment model architecture. In this paper, we use the state-of-the-art alignment model \citep{xu2019cross} as our baseline method and propose two ways to enhance this model by incorporating easy assignment information.

\subsubsection{Alignment Model Baseline.}
\citet{xu2019cross} utilized a graph (namely \textit{topic graph}) to capture the context information of an entity 
(namely \textit{topic entity}) within the KG. 
For instance, Figure~\ref{fig:framework} gives the topic graphs of George Bush in both the Chinese and English KG.
The entity alignment task is then viewed as a graph matching problem, whose goal is to calculate the similarity of these two topic graphs, say $G_{1}$ and $G_{2}$.
To achieve this, they further propose a neural graph matching model that includes the following four layers:
\begin{itemize}
    \item \textbf{Input Representation Layer.} The goal of this layer is to learn embeddings for entities that occurred in topic entity graphs by using a graph convolution neural network (GCN) \citep{kipf2016semi}.
    \item \textbf{Node-Level Matching Layer.} This layer is designed to capture local matching information by comparing each entity embedding of one topic entity graph against all entity embeddings of the other graph in both ways (from $G_{1}$ to $G_{2}$ and $G_{2}$ to $G_{1}$).
    \item \textbf{Graph-Level Matching Layer.} In this layer, the model applies another GCN to propagate the local matching information throughout the graph. The motivation behind it is that this GCN layer can encode the global matching state between the pairs of whole graphs. The model then feeds these matching representations to a fully-connected neural network and applies the element-wise \textit{max} and \textit{mean} pooling method to generate a fixed-length graph matching representation.
    \item \textbf{Prediction Layer.} The model finally uses a two-layer feed-forward neural network to consume the fixed-length graph matching representation and applies the \textit{softmax} function in the output layer.
\end{itemize}

\subsubsection{Our Model.}
In contrast to \citet{xu2019cross} that only takes two topic graphs as input, we can utilize additional information such as easy assignments found in previous decoding steps to resolve hard assignments.
In particular, we introduce two ways to enhance this baseline model by explicitly integrating the easy assignment information into two layers of \citet{xu2019cross}:
\begin{itemize}
    \item \textbf{Enhanced Input Representation Layer.} In this layer, \citet{xu2019cross} utilizes a GCN to learn entity embeddings from the topic graph, where the entity surface form has been proved to be a key feature in deriving their embeddings. Therefore, we require that the aligned entities found in the easy alignments should have the same surface forms so that they could share the common embeddings. For example, in Figure~\ref{fig:framework}, after the first round of decoding, {\small{哈佛大学}} (Harvard University) is aligned to Harvard, we then change the surface form of ``{\small{哈佛大学}}'' to ``Harvard'' in the second decoding step.
    \item \textbf{Enhanced Node-Level Matching Layer.} As concluded in \cite{xu2019cross}, the node-level matching layer has a significant impact on the matching performance, since it captures the local entity matching information. In the baseline model, the entity similarities are calculated based on the entity embeddings derived from the first GCN layer. Although in the enhanced input representation layer the aligned entities have the same surface forms, it can still not guarantee that their embeddings are close. It is because the first GCN layer is supposed to encode not only the surface form but also the structural information into their representations. Therefore, we explicitly incorporate the easy alignment information into this layer by enforcing that the normalized similarities between the aligned entities to be $1.0$. Then, we feed the revised entity similarities to the graph-level matching layer and the final prediction layer.
\end{itemize}
Notice that, in practice, there are two possible options to build the enhanced alignment model in our framework.
First, we can directly use the pre-trained baseline but replace its first two layers with our proposed enhanced layers.
Because we do not modify the model architecture, no more parameters are needed to be learned.
The second way is to train a new enhanced alignment model with randomly sampled alignments as simulated easy alignments.
The motivation behind is that given more easy alignments, the model could more focus on learning to disambiguate hard alignments.
Experimental results show that the latter achieves much better performance.
We will discuss these two options in the experiment section.

\subsection{Joint Entity Alignment}
\begin{figure}[t!]
    \centering
    \includegraphics[width=1.0\linewidth]{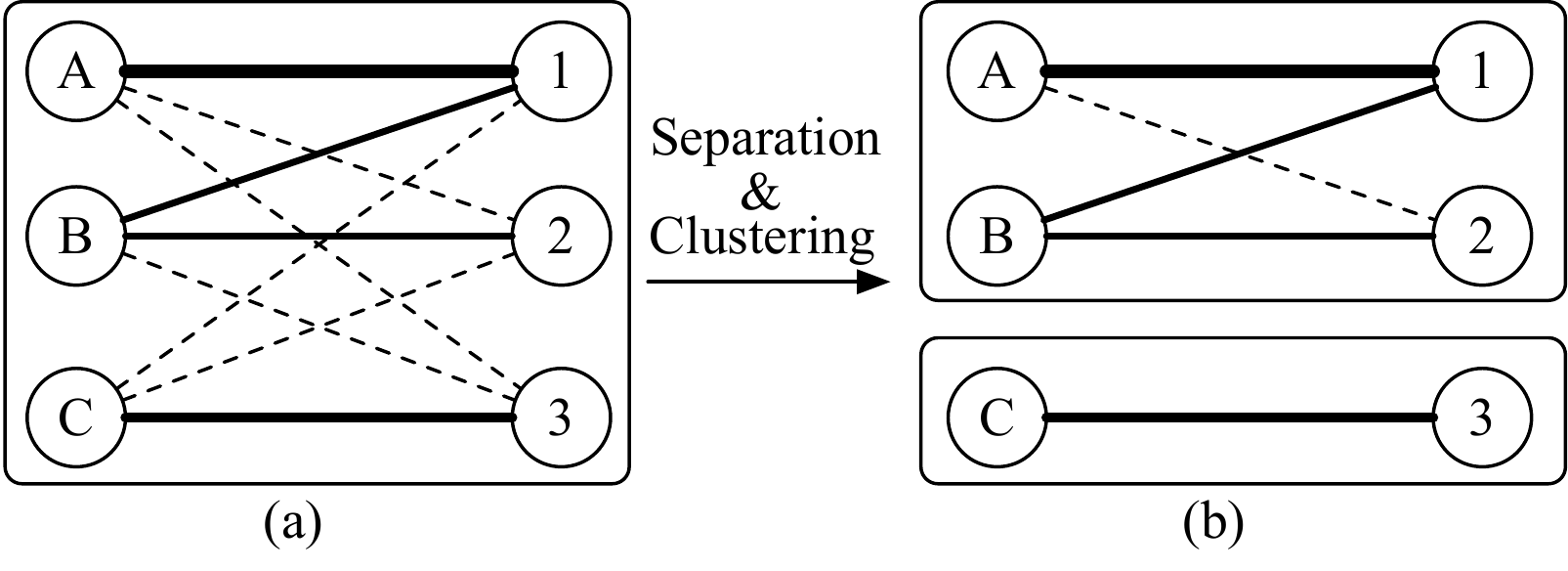}
    \caption{(a) The original alignment results between entities \{$A$, $B$, $C$\} $\in G_{1}$ and \{$1$, $2$, $3$\} $\in G_{2}$, where the thickness of a line represents its alignment probability and very weak alignments are shown as dotted lines;
    (b) The sub-spaces after separation.}
    \label{fig:space_split}
    \vspace{-3mm}
\end{figure}

As shown in Figure~\ref{fig:space_split}(a), our model typically outputs a 2-dimensional matrix of probabilities after decoding, where each cell item (such as $p(e_t|e_s)$) represents the likelihood of aligning source entity $e_s$ to target entity $e_t$.
The goal of the entity alignment task is then equal to find the best solution (a set of one-to-one alignments) with the highest probability:
\begin{equation} \label{eq:ori_goal}
    \argmax_{\mathbf{A}} \prod_{(e_s, e_t) \in \mathbf{A}} p(e_t|e_s) 
    % \textrm{,}
\end{equation}
where $\mathbf{A}$ represents one solution.
Since knowledge graphs are usually huge, this problem cannot be solved by naive enumeration, which takes $O(N!)$ time for KGs with $N$ entities.
Existing works choose the optimal local match for each source entity while neglecting the \textit{one-to-one} nature, and as a result, multiple source entities may be mapped to one target entity.

Here, for the first time, we propose to explicitly incorporate this one-to-one constraint into the alignment prediction.
To achieve this, we first reformat the goal from maximizing the product of probabilities (Equation \ref{eq:ori_goal}) to minimizing the sum of negative log-likelihoods.
\begin{equation}
    \argmin_{\mathbf{A}} \sum_{(e_s, e_t) \in \mathbf{A}} -\log p(e_t|e_s)
\end{equation}
As a result, the entity alignment problem is equivalently converted to the well-studied ``task assignment'' problem\footnote{https://en.wikipedia.org/wiki/Assignment\_problem}, where each agent/task is assigned to exactly one task/agent, and each agent-task assignment has a fixed cost that does not depend on the other assignments.

The Hungarian algorithm \citep{kuhn1955hungarian} has been proven to be efficient for finding the best solution for this problem.
It takes a cost matrix as input, which can be easily achieved by padding rows or columns of a constant value for the non-square matrix.
For a brief introduction, the algorithm takes the following four main steps for the cost matrix with $N\times N$ elements, where the last two steps repeat until a solution is found.\footnote{http://www.hungarianalgorithm.com/ provides a detailed explanation and an online demo.}
It is guaranteed that a solution could be found within $O(N^4)$ time.

\vspace{0.2em}
\begin{center}
\begin{tabularx}{0.95\linewidth}{cX}
    \toprule
    Step & Description \\
    \midrule
    1 & Find the lowest item for each row and subtract it from the others in that row. \\
    2 & Similarly, find the lowest item for each column and subtract it from the others in that column. \\
    3 & Cover all zeros in the resulting matrix using a minimum number of horizontal and vertical lines. If \emph{less than} $N$ lines are required, go to Step 4; otherwise, a solution is found. \\
    4 & Find the smallest item $v$ not covered by any line in Step 3. Subtract $v$ from all uncovered items, and add $v$ to all items covered by two lines. Go to Step 3. \\
    \bottomrule
\end{tabularx}
\end{center}
\vspace{0.2em}
One can see that naively applying Hungarian is impractical, as it still takes $O(N^4)$ computation time for matching two KGs of $N$ nodes.
To further decrease the time consumption, we break the whole search space into many isolated sub-spaces, where each sub-space contains only a subset of source and target entities for making alignments.
Specifically, we discard the candidate alignments with a probability lower than a predefined threshold $\tau$ from the original search space.
Based on this, we define two source entities being connected only if they share common candidates in the target.
Doing in this way fits the intuition where a large KG usually contains many domains, such as politics, sports and science, and only the entities within each domain have densely interacted.
Our experiments show that $\tau$ has little effect on performance, while it dramatically reduces the search time.

Figure \ref{fig:space_split} illustrates the search space separation process, where thin and dotted lines correspond to low-confident alignments.
After dropping out these alignments with low model scores, the whole search space is split into two independent sub-spaces, as shown in Figure~\ref{fig:space_split}(b).
Here \textcircled{\small{A}} and\textcircled{\small{B}} are in the same sub-space, as they share the same target candidate \textcircled{\small{1}}.
Removed connections (such as \textcircled{\small{A}} to \textcircled{\small{2}}) are considered as infinite cost.
As the next step, each sub-space is solved with the Hungarian algorithm, before their results are combined to form our final outputs.

\begin{table}[t!]
\centering
\small
\begin{tabular}{c|c|c|c|c}
\toprule[0.8pt]
% \multirow{2}{*}{Method} & \multicolumn{2}{c|}{\textit{ZH}-\textit{EN}} & \multicolumn{2}{c|}{\textit{EN}-\textit{ZH}} & \multicolumn{2}{c|}{\textit{JA}-\textit{EN}} & \multicolumn{2}{c|}{\textit{EN}-\textit{JA}} & \multicolumn{2}{c|}{\textit{FR}-\textit{EN}} & \multicolumn{2}{c}{\textit{EN}-\textit{FR}} \\
\multicolumn{2}{c|}{\textbf{Datasets}} & \textbf{Entities} & \textbf{Relations} & \textbf{Triples} \\
\hline
\multirow{2}{*}{DBP15K$_{\textit{ZH}-\textit{EN}}$} & Chinese & 66,469 & 2,830 & 153,929 \\
& English & 98,125 & 2,317 & 237,674 \\
\hline
\multirow{2}{*}{DBP15K$_{\textit{JA}-\textit{EN}}$} & Japanese & 65,744 & 2,043 & 164,373 \\
& English & 95,680 & 2,096 & 233,319 \\
\hline
\multirow{2}{*}{DBP15K$_{\textit{FR}-\textit{EN}}$} & French & 66,858 & 1,379 & 192,191 \\
& English & 105,889 & 2,209 & 278,590 \\
\toprule[0.8pt]
\end{tabular}
\caption{Dataset summary.}
\label{tab:dataset_sum}
\end{table}
\section{Experimental Setup}
\paragraph{Datasets.}
We evaluate our approach on three large-scale cross-lingual datasets from DBP15K \citep{sun2018cross}.
These datasets are built upon Chinese, English, Japanese and French versions of DBpedia \citep{auer2007dbpedia}.
Each dataset contains 15,000 inter-language links connecting equivalent entities in two KGs of different languages. We use the same training/testing split as previous works, $30$\% for training, and $70$\% for testing.
Table~\ref{tab:dataset_sum} lists their statistical summaries.\\\\
\textbf{Evaluation Metrics.}
Like previous works, we use \textit{Hits}@1 to evaluate our model, where a \textit{Hits}@$1$ score (higher is better) is computed by measuring the proportion of correctly aligned entities ranked in the top one.\\\\
\textbf{Comparison Models.}
We compare our approach against existing alignment methods: JE \citep{hao2016joint}, MTransE \citep{chen2017multilingual}, JAPE \citep{sun2018cross}, IPTransE \citep{zhu2017iterative}, BootEA \citep{sun2018cross}, GCN \citep{wang2018cross}, GM \citep{xu2019cross} and RDGCN \citep{wu2019relation}.\\\\
\textbf{Model Variants.}
To evaluate different reasoning methods,
we provide three implementation variants of our model
for ablation studies, including
(1) \textit{X}-EHD: the baseline model \textit{X}
that only uses our proposed \textbf{E}asy-to-\textbf{H}ard \textbf{D}ecoding strategy;
(2) \textit{X}-JEA: the baseline model \textit{X} that only uses our proposed \textbf{J}oint \textbf{E}ntity \textbf{A}lignment method; (3) \textit{X}-EHD-JEA: the baseline model \textit{X} that uses both of these two reasoning methods.\\\\
\textbf{Implementation details.}
For the configurations of the alignment model, we use the same settings as \cite{xu2019cross}. Specifically, we use the Adam optimizer \citep{DBLP:journals/corr/KingmaB14} to update parameters with mini-batch size 32.
The learning rate is set to 0.001.
The hop size of two GCN layers is set to 2 and 3, respectively.
Following \cite{wu2019relation}, we use Google Translate to translate Chinese, Japanese, and French entity names into English, and then use Glove embeddings \citep{pennington2014glove} to construct the initial entity representations in the model.
For all datasets, we first use the baseline model to retrieve the top $10$ alignments, normalize their scores as probabilities and then perform the proposed coordinated reasoning methods over them.
For the Easy-to-Hard decoding method, $\alpha$ is set to $0.75$, and $K$ is set to 20. For the joint entity alignment, $\tau$ is set to 0.10.
For training the enhanced alignment model, for each topic graph pair, we randomly choose at most two gold alignments from the ground truth as the simulated easy alignments.

\section{Results and Discussion}
\begin{table}[t!]
\centering
\small
\begin{tabular}{l|c|c|c}
\toprule[0.8pt]
% \multirow{2}{*}{Method} & \multicolumn{2}{c|}{\textit{ZH}-\textit{EN}} & \multicolumn{2}{c|}{\textit{JA}-\textit{EN}} & \multicolumn{2}{c}{\textit{FR}-\textit{EN}} \\
Method & \textit{ZH}-\textit{EN} & \textit{JA}-\textit{EN} & \textit{FR}-\textit{EN} \\
% \cline{2-7}
% & @1 & @10 & @1 & @10 & @1 & @10 \\
\hline
% \newcite{hao2016joint}
JE & 21.27  & 18.92  & 15.38  \\
% \newcite{chen2016multilingual} 
MTransE & 30.83 & 27.86 & 24.41 \\
% \newcite{sun2017cross} 
JAPE & 41.18 & 36.25 & 32.39 \\
IPTransE & 40.59 & 36.69 & 33.30 \\
% \newcite{wang2018cross} 
GCN & 41.25 & 39.9 & 37.29 \\
BootEA & 62.94 & 62.23 & 65.30 \\
GM & 67.93 & 73.97 & 89.38 \\
RDGCN & 70.75 & 76.74 & 88.64 \\
\hline
\hline
GCN -JEA & 43.43 & 45.00 & 39.78 \\
BootEA-JEA & 64.56 & 64.17 & 69.31  \\
RDGCN-JEA & 72.03 & 77.56 & 90.49 \\
\hline
\hline
GCN-EHD & 44.37 & 41.72 & 39.09 \\
BootEA-EHD & 65.27 & 65.36 & 68.92 \\
RDGCN-EHD & 71.15 & 77.07 & 91.01 \\
\hline
\hline
GM-EHD & 70.31 & 77.92 & 90.49 \\
GM-JEA & 72.05 & 78.73 & 91.08 \\
GM-EHD-JEA & \textbf{73.58} & \textbf{79.15} & \textbf{92.43} \\
\toprule[0.8pt]
\end{tabular}
\caption{Evaluation results on the datasets.}
\label{tab:results}
\end{table}
\subsection{Main Results}
Table~\ref{tab:results} shows the performance of all compared approaches on the evaluation datasets.
We can see that both of the Easy-to-Hard decoding strategy (referred as EHD in Table~\ref{tab:results}) and the joint entity alignment method (referred as JEA in Table~\ref{tab:results}) could significantly improve the performance of GM.
When these two methods are combined, the overall performance is further improved, outperforming previous works.
We also investigate whether our proposed reasoning methods could also boost existing baselines.
From Table~\ref{tab:results}, we can see also that the joint entity alignment method could also improve the performance of GCN, BootEA and RDGCN, indicating that
our method is able to avoid the \textit{many-to-one} problem effectively.
Recall that, the Easy-to-Hard decoding method requires an enhanced alignment model that could integrate the easy alignment information.
Since designing enhanced versions for these baselines is beyond our goal, here we only enforce that
that the aligned entities found in the easy alignments have the same surface form.
We find that this simplified strategy could still improve these baselines, which also suggests that our proposed decoding strategy is generally helpful to the alignment models.

\subsection{Discussion}
Let us first look at the impacts of alignment-dropping threshold $\tau$ to both the performance and running time for our joint entity alignment algorithm.
From Table \ref{tab:dev_tau}, we can see that
decreasing $\tau$ can slightly improve the performance but with a huge cost of computation time.
For example, when $\tau$ is changed from $0.15$ to $0.10$, the accuracy could increase by $0.12$\% but the computation time dramatically increases from $39$s to almost $25$ minutes.
Moreover, if $\tau$ is set to $0.05$, we cannot even get the results.
As shown in Table~\ref{tab:dev_tau},
in order to better understand why the running time changes,
we additionally analyze the size of the largest sub-space.
We find that the size of the maximal sub-space under $\tau=0.05$ is $3$ times more than the size under $\tau=0.10$, thus the running time under $\tau=0.05$ is expected to be roughly $32$ hours, which is $81$ ($3^4$) times than the time under $\tau=0.10$.
The running time does not significantly change when increasing $\tau$ from 0.15 to 0.20, because the Hungarian algorithm does not take much time for this situation, and the most time consumption is data processing.
\begin{table}[t!]
    \centering
    \begin{tabular}{c|c|c|c}
    $\tau$ & Max sub-space & Time & \textit{FR-EN}(hit@1) \\
    \hline
      0.05 & 5238 & -- & -- \\ % 48661s ==> 13h31m, 89.07
      0.10 & 1562 & 24m34s & 91.02 \\
      0.15 & 116 & 39s & 90.90 \\
      0.20 & $<$100 & 38s & 90.78 \\
    \end{tabular}
    \caption{Performance and computation time for different $\tau$ values, where \emph{Max sub-space} shows the number of source nodes in the largest sub-space. Baseline accuracy is 89.38.}
    \label{tab:dev_tau}
\end{table}

\begin{table}[t!]
    \centering
    \begin{tabular}{c|c|c|c|c}
    $\alpha$ & Decoding Rounds & \textit{ZH-EN} & \textit{JA-EN} & \textit{FR-EN} \\
    \hline
     -- & -- & 66.29 & 72.31 & 88.07 \\
      0.95 & 2 & 69.05 & 74.33 & 88.25 \\ % 48661s ==> 13h31m, 89.07
      0.85 & 5 & 71.71 & 75.10 & 90.31 \\
      0.75 & 10 & \textbf{72.09}  & \textbf{76.62} & \textbf{91.18}  \\
      0.65 & 20 & 67.12 & 72.15 & 88.60  \\
    \end{tabular}
    \caption{Hit@1 accuracies and decoding rounds on the development set for different $\alpha$ values. The first row lists the accuracies of the GM baseline.}
    \label{tab:dev_alpha}
\end{table}

% Next, we will have a closer look at the effect of our Easy-to-Hard decoding method.
We also investigated the impact of the probability threshold $\alpha$ on the performance for our Easy-to-Hard decoding method.
We experimented with different $\alpha$ values and evaluated our model on the development set of the DBP15K.
Table~\ref{tab:dev_alpha} reports hit@1 accuracies on these datasets.
We can see that our model could benefit from decreasing $\alpha$ until it reaches $0.75$.
It is expected to find that lower $\alpha$ may hurt the performance since it incorporates some incorrect predictions as easy (gold) alignments into the model.
Recall that in our decoding algorithm, we continuously perform the inference until less than $K$ new easy alignments are found in the previous round.
As shown in Table~\ref{tab:dev_alpha}, we observed that decreasing $\alpha$ not only achieves worse performance but also requires more converge rounds.
To better understand why more converge rounds are required, we analyzed the intermediate established alignments during the inference.
We find this is due to those incorrect alignments introduced by reducing $\alpha$ produce a chain reaction, which offers the model more confidence about some uncertain but incorrect alignments, resulting in more decoding rounds.

Recall that there are two options to build the enhanced alignment model, where the first one directly replaces two layers of a pre-trained GM model with our proposed enhanced layers while keeping the parameters the same; the second one trains a new GM model with simulated easy alignments.
We evaluate these two options on several datasets and observe that both of these two ways could improve the performance but the model could gain more performance improvement from the second way.
We further manually analyze the predicted alignments of these two options and find that the new trained GM model could resolve more ambiguous (hard) alignments.
We think this is due to that introducing the simulated easy alignments into the training phase could allow the model to learn how to properly utilize these additional evidence to disambiguate the hard alignments.

Here one natural question is how many simulated easy alignments are needed for training the new GM model.
In experiments, we find that using two simulated easy alignments to train the model could achieve the best performance; introducing more easy alignments to train the model could not further improve the results.
However, this observation is in conflict with our intuition, that is, more easy alignment information could better help the model to disambiguate those uncertain predictions.
By analyzing the entities in the test set, we find this is due to that among these entities, at most three entities co-occur in the same topic graphs,
and consequently, during the decoding, the model could only introduce at most two easy alignments.
Motivated by this observation, we conducted an additional experiment that predicts alignments for all entities in the KGs except the training seeds. We find that our reasoning methods could achieve more performance improvement, and considering more than two easy alignments into the training also further improves the overall performance as we expected.
Note that, although this experiment may consume almost $5$ times more than the original decoding time, we believe that some optimization could be adopted to reduce the time complexity, which we leave for the future work.

% many of these entities are not in the same topic graphs.

% Consequently.

\section{Conclusion}
Previous entity alignment methods mainly use the same decoding strategy that independently chooses the optimal local match for each source entity without considering the global alignment coherence, thereby may cause the many-to-one problem. To address this, we propose two reasoning method, including a new Easy-to-Hard decoding strategy and joint entity alignment method.
Specifically, the Easy-to-Hard decoding method iteratively decodes the test set by taking the most model-confident alignments predicted in the previous iteration as additional inputs to the current iteration for resolving the model-uncertain alignments.
The joint entity alignment method views the entity alignment as the task assignment problem and employs the Hungarian algorithm to guarantee the predicted alignments are one-to-one mappings.
Experimental results on the DBP15K dataset show that our reasoning methods are general to these baselines and can significantly improve their performance.

\begin{quote}
\begin{small}
\bibliographystyle{aaai}
\bibliography{aaai}
\end{small}
\end{quote}

\end{CJK*}
\end{document}